\documentclass[conference]{IEEEtran}
\usepackage{graphicx} 
\usepackage{subfig}
\usepackage{url}
\usepackage[hmargin=1.9cm,vmargin=1.9cm,letterpaper]{geometry} 

\title{\ \\ Effects of Compensation, Connectivity and Tau in a Computational Model
of Alzheimer's Disease.} 

\author{Mark Rowan, \url{m.s.rowan@cs.bham.ac.uk}\\ \textit{School of Computer
Science, University of Birmingham, B15 2TT, UK}}

\begin{document}
\maketitle

\begin{abstract}
This work updates an existing, simplistic computational model of
Alzheimer's Disease (AD) to investigate the behaviour of
synaptic compensatory mechanisms in neural networks with small-world
connectivity, and varying methods of calculating compensation. It
additionally introduces a method for simulating tau neurofibrillary pathology,
resulting in a more dramatic damage profile. Small-world connectivity is shown
to have contrasting effects on capacity, retrieval time, and robustness to
damage, whilst the use of more easily-obtained remote memories rather than
recent memories for synaptic compensation is found to lead to rapid
network damage.
\end{abstract}

\section{Introduction}
Alzheimer's disease (AD) is a specific form of dementia, characterised
biologically by neurofibrillary tau protein tangles and beta-amyloid (A$\beta$)
protein plaques \cite{tiraboschi2004}, and symptomatically by a progressive
decline in memory capabilities. In particular, recent memories are the first to
be lost whilst distant memories are retained, but as the disease progresses this
is followed by gradual total loss of recall, a corresponding loss of
personality, motor control, and other bodily functions, and finally death
\cite{francis1999cholinergic}.

Computational modelling of neurological disorders such as AD is an established
tool \cite{aakerlund1998neural} but existing models of AD such as
\cite{hasselmo1994runaway, ruppin1995neuralimpairment} can now be improved in
line with better understanding of the disease. One such model, by Ruppin and
Reggia (1995) \cite{ruppin1995neuralimpairment}, showed how simple lesions in a
single-layer associative network trained in an \textit{activity-dependent}
Hebbian manner leads to loss of memory, and the addition of a local compensation
factor causes the pattern of functional damage to mirror more closely that found
in AD whereby recently-stored memories are lost before historical memories.
Later work showed how the compensation factor can be made biologically plausible
by depending only on the post-synaptic potential of the remaining neurons after
lesioning \cite{horn1996neuronal}.

This model remains widely cited \cite{duch2007computational,
savioz2009contribution} even though it could be made capable of better
approximation of the lesions representing AD pathology; currently it only either
deletes neurons and synapses at random or deletes neurons within a specified
radius on a 2-D grid \cite{ruppin1995patterns}. Today we know much more about
connectivity strategies in the brain such as small-world networks
\cite{watts1998collective, bullmore2009complex} as well as the biological
processes underpinning AD, such as neurofibrillary tau pathology.

In this paper, methods are presented for enhancing the Ruppin and Reggia model
with up-to-date techniques which may be more representative of the underlying
biology. This work is intended to examine differences in behaviour which may
occur when considering connectivity strategies, specific details of compensatory
techniques and lesioning in accordance with specific pathologies, with the aim
of leading to development of more accurate representations of a range of
pathological processes underlying AD such as those involving tau, beta-amyloid,
and N-amyloid precursor protein \cite{nikolaev2009napp}, in more complex network
models such as LEABRA \cite{oreilly2001generalization}, spiking neural networks
\cite{gerstner2002spiking}, and reservoir networks
\cite{lukosevicius2009reservoir}.

The remainder of this paper is organised as follows: Section \ref{model}
describes the Ruppin and Reggia model in greater detail and the updates made to
it in this work, section \ref{results} presents the results of experiments
characterising the network's behaviour with these new enhancements, and section
\ref{conclusions} deals with concluding remarks and outlines future directions in
which this research could be taken.

\section{Model Description}
\label{model}
\subsection{Learning rule}
Ruppin and Reggia showed how a variant of an attractor network model proposed by
Tsodyks and Feigel'Man (the T-F model) \cite{tsodyks1988enhanced} is capable of
storing patterns in a biologically-plausible Hebbian \textit{activity-dependent}
manner. This is achieved using a repetitive-learning process whereby each
pattern to be stored ``must be presented to the network several times before it
becomes engraved on the synaptic matrix with sufficient strength, and is not
simply enforced on the network in a `one-shot' learning process''
\cite{ruppin1995neuralimpairment}. An updated version of the model
\cite{ruppin1995patterns} added Gaussian partial-connection of the network
rather than full connectivity.

\begin{equation}
\label{eq:actdep}
W_{ij}(t) = W_{ij}(t-1) + \frac{\gamma}{N}(S_{i}-p)(S_{j}-p) 
\end{equation}

The network learns patterns through a process of activity-dependent learning
according to the update rule in equation \ref{eq:actdep}. A set of
\textit{external inputs} delivers activation greater than the neural threshold
to each unit of the network according to the pattern to be learned. $W$ is the
weight matrix of undirected connections between neurons $i$ and $j$, $\gamma$ is
a constant determining the magnitude of activity-dependent changes, $N$ is the
number of neurons in the network, $K \leq N$ is the number of other units to
which each unit is connected, $S$ refers to the neuronal state $\{0,1\}$, and
$p$ is the coding rate denoting the proportion of \textit{1}s compared to
\textit{0}s in the stored memory patterns ($p \ll 1$ as cortical networks are
found to have low coding rates \cite{abeles1990firing}).

The activity-dependent learning rule for pattern storage is based on the Hebbian
principle but introduces the requirement for each given pair of units to remain
in the same state for a certain number of update iterations (the suggested value
is 5) before the synaptic weight between them is updated, and requires each
pattern to be presented several times in turn to the network before it is
completely stored. Thus the learning algorithm attempts to mitigate the effects
of the Hebb rule's ability to globally alter synaptic weights in a
biologically-unrealistic way and circumvents its method of storing each pattern
in a `one-shot' process which is susceptible to the presence of errors or noise.
By presenting each pattern several times to the network, any noise present in
the inputs is reduced and the synaptic matrix is gradually constructed rather
than being enforced in a single process by the learning rule.

\subsection{Performance evaluation}
Patterns are recalled using a noisy version of the complete pattern applied to
the network via the same set of external inputs used for learning with
activation less than the neural firing threshold. A measure of the recall
performance the network for a given pattern $\xi^\mu$, termed the
\textit{overlap} between the resulting network state and the pattern, has the
effect of counting the correctly-firing units whilst also penalising with a
lower weighting those units which fire erroneously (equation \ref{eq:overlap})
\cite{tsodyks1988enhanced}:

\begin{equation}
\label{eq:overlap}
m^{\mu} (t) = \frac{1}{p (1 - p) N} \sum_{i = 1}^{N} (\xi ^{\mu} _{i} - p) S_i
(t)
\end{equation}

\subsection{Synaptic compensation}
In the work by Ruppin and Reggia the network model was lesioned by deleting
synapses or neurons at random and implementing a process of \textit{variable
synaptic compensation}, where ``the magnitude of the remaining synapses is
uniformly strengthened in a manner that partially compensates for the decrease
in the neuron's input field'' \cite{ruppin1995neuralimpairment} by multiplying
the weights of the remaining synaptic connections by a globally-determined (i.e.
depending on knowledge of the overall fraction of deletion) local compensation
factor.

Ruppin and Reggia examined the overall degradation in recall performance and the
pattern of relative sparing of older memories compared to recently stored
patterns (as observed in AD patients \cite{kopelman1989remote}) as the network was
progressively lesioned, and concluded that synaptic deletion and compensation in
this model can be demonstrated to reveal similar symptoms to the cognitive
decline observed in AD.

However a global synaptic compensation strategy is biologically implausible as
each neuron must somehow be aware of the global deletion rate both for itself,
and for other neurons around it. Horn et al. \cite{horn1996neuronal} therefore
introduce a neuronal-level compensatory mechanism which causes each neuron to
adjust its output based only on changes in the neuron's average post-synaptic
potential (or summed input), and which does not require the explicit knowledge
of either global or local levels of synaptic deletion.

At any given moment, each neuron has an estimate $\hat{w}_i$ of its total
connectivity compared to the starting value $(w_i = 1)$. It can compensate for
this reduced connectivity by multiplying the remaining incoming synapses
(essentially, lowering its firing threshold) by a value $c_i$. This is achieved
via repetition of the following steps:

\begin{itemize}
  \item In the pre-morbid state (i.e. before each iteration of lesioning) a
  set of random noise patterns $(p \ll 1)$ is presented to the network and it is
  allowed to fall into a stable state. Each neuron then obtains its resulting
  input field measurement, the expected value of which is denoted $\langle h_i
  ^2 \rangle$.
  \item Horn et al. state that $\langle h_i^2 (\hat{w}_i) \rangle = c_i^2
  \hat{w}_i \langle h_i^2 (w_i = 1) \rangle$. Given an assumption that $c^2_i
  \hat{w}_i = 1$ (i.e. the network is currently correctly compensating for any
  value of $w<1$), the neuron's average ``noise-state'' input field value, the
  expected value of which is denoted $\langle R_i^2 \rangle$, is therefore
  equivalent to $\langle h_i^2 \rangle$.
  \item The same process is repeated using a set of already-stored patterns
  rather than random noise patterns. Each neuron then obtains its resulting
  average ``signal-state'' input field strength, the expected value of which is
  denoted $\langle S_i ^2 \rangle$ (Horn et al. speculate that this process could occur
  biologically during dreaming). As with the earlier noise term, $\langle S_i ^2
  \rangle \equiv \langle h_i ^2 \rangle$ as $c_i^2 \hat{w}_i = 1$.
  \item The network is lesioned in some way unknown to the individual neurons
  (e.g. by deleting synapses).
  \item Now, in order to estimate the new value of $\hat{w'}_i$ in the
  post-morbid state, and thus to compute a new value for $c'_i$, a further set
  of already-stored patterns is presented to the network and the network allowed to
  converge once more to a stable state. A new \textit{post-morbid} value for each
  neuron's input field $\langle h_i ^2 \rangle$ is obtained.
  \item Horn et al. separate this $\langle h_i ^2 \rangle$ into signal and noise
  terms: $\langle h_i ^2 (w_i) \rangle = c_i ^2 \hat{w}_i^2 \langle S_i ^2
  \rangle + c_i^2 \hat{w}_i \langle R_i ^2 \rangle$. The noise term is already
  known from the earlier steps, and is subtracted from the post-morbid input
  field value. It is thus possible to calculate $\hat{w'}_i^2$ using equation
  \ref{compensation}, and then to derive a new value for the compensation
  $c'_i$:
\end{itemize}

\begin{equation}
\hat{w'}_i^2 = \frac{\langle h_i ^2 \rangle - \langle R_i ^2 \rangle}{c_i^2
\langle S_i ^2 \rangle}
\label{compensation}
\end{equation}

\subsection{Unanswered questions}
Whilst experimental support exists for the predicted compensatory strengthening
of synapses in AD \cite{savioz2009contribution}, one unexplained result drawn
from this model is that significant neuronal deletion (around $50\%$) in the
model is required before memory function is seriously impaired. This rate of
deletion is much larger than the rate observed clinically in the latest stages
of Alzheimer's disease (between $10\%$ and $30\%$ reduction of volume in the
hippocampal regions in severe cases of AD \cite{minati2009reviews} and certainly
far more than the $10\%$ general cerebral atrophy reported at initial diagnosis
of the disease \cite{ridha2006tracking}), implying that there must be other
factors additionally affecting cognitive decline.

A further limitation of the model is that lesioning is performed only by
deleting a number of randomly-selected neurons or connections at each step,
which does not necessarily represent the subtleties of the underlying pathology.
The authors present a method of applying lesions in a localised spatial manner
by deleting all of the neurons and/or connections within a circle or rectangle
of a given area \cite{ruppin1995patterns}, but this does not incorporate any of
the known neurodegenerative mechanisms such as tau or amyloid pathology.

During the synaptic compensation process previous studies have not examined the
differences in performance when using recent versus remote memories to calculate
the signal term.

Finally, the method in which the network is interconnected (either fully, or
using an arbitrary number of connections per neuron in a localised Gaussian
manner) is again simplistic and does not represent biologically realistic
connection strategies such as small-world networks \cite{watts1998collective} or
neural Darwinism (pruning of weaker synapses during development)
\cite{hoffman2001neural}.

\subsection{Implementing different connectivity strategies}
Biological neural networks such as those found in the hippocampus are generally
sparsely connected \cite{levy1996sequence, bullmore2009complex}. It has been
shown that in Alzheimer's disease, small-world clustering (as measured by the
clustering coefficient) is significantly reduced at a global level, resulting in
large changes to the local organisation of the network \cite{supekar2008network}.

Connection dilution mechanisms for associative networks include connecting each
unit over a flat random distribution, wiring each unit in a spatial manner to
those immediately surrounding it with Gaussian probability, and using a
randomised small-world network connection strategy \cite{watts1998collective}.

Small-world networks in this model are constructed in the form prescribed by
Watts and Strogatz \cite{watts1998collective} by firstly connecting each neuron
to its closest $K$ neighbours. Then, according to a probability of re-wiring
$p(rewire)$, the connections between each unit and its two immediate neighbours
are randomly assigned to other units in the network. Once each unit in the
network has been considered, the neighbours two places away from each unit are
then considered, and then those three places away, until each connection in the
network has finally been randomly re-wired or left in place.

It has been shown that a process of diluting the synaptic weight matrix of a
Hopfield network such that it is no longer fully-connected still causes it to
behave in much the same manner as a fully-connected network when stored state
vectors are generally of low activity \cite{evans1989random}.

\subsection{Implementing tau lesioning}
\label{tauimplementation}
\subsubsection{Medical background}
The tau hypothesis refers to the neurodegenerative effects of a modified (or
\textit{hyperphosphorylated}) form of the tau protein which aggregates with
other fibres of tau and eventually forms the neurofibrillary tangles (NFTs)
inside neurons which are prevalent in brains with AD. It proposes that the
cognitive decline in AD is due primarily to loss of synapses and neurons (via a
toxic form of the modified tau), and the subsequent loss of connectivity experienced
\cite{spires2009tau}.

The normal function of tau is disrupted in AD: ``tau is essential for
establishing neuronal cell polarity and axonal outgrowth during development and
for maintaining axonal morphology and axonal transport [of
neurotransmitter-containing vesicles along the axon] in mature cells''
\cite{feinstein2005inability, johnson2004tau} and in both constructing and
stabilising microtubules which, in developing neurons, are important for
establishing neuronal cell polarity and outgrowth, and in adult neurons are
essential for proper structure, function and viability
\cite{feinstein2005inability}. Instead of binding to the microtubules, tau in AD
becomes sequestered into NFTs within the neurons \cite{ballatore2007tau} and as
the level of normal tau in the brain is reduced the microtubules disintegrate,
causing further neuronal dysfunction. The existence of NFTs could also present a
toxic gain-of-function by physically obstructing the transport of vesicles
within the neuron (leading to cognitive impairment) and also by further
sequestration of normal tau into the modified form as part of a cascade of
neurodegeneration \cite{ballatore2007tau}.

Although the amyloid hypothesis provides a possibly more widely encompassing
view of AD, it has a number of significant unexplained problems, not least that
``the number of amyloid deposits in the brain does not correlate well with the
degree of cognitive impairment'' \cite{hardy2002amyloid}, and so the tau
hypothesis and its relationship to the amyloid hypothesis remain an important
subject for further research.

\subsubsection{Computational implementation}
Whilst it is possible to delete either neurons or the connections between them
as shown in earlier studies \cite{ruppin1995neuralimpairment}, these processes
can be considered essentially the same: if all the incoming synapses of a 
particular neuron are deleted, the neuron is no longer able either to receive
activation from surrounding neurons, nor is it able to have any excitatory effect
on its neighbours. This neuron might just as easily be considered to have been
deleted, as it is effectively removed from the network completely.

Hopfield-type networks (including the T-F network) also suffer from the inherent
problem that the output layer is essentially the only layer of the network. That 
is, whenever lesions are applied to the network by deleting neurons, this
necessarily results in the inability of the network to completely recover a cued
pattern regardless of the effects the removal of these neurons may have had on
the underlying pathology of the network, as the `output layer' is now only
partially complete and can no longer map with full accuracy to every given 
pattern, resulting in a perceived decrease in network performance.

So it is suggested that results more representative of the underlying pathology
could be obtained by the introduction of a subtle shift from \textit{synaptic 
loss} to \textit{neuronal atrophy}, whereby whole groups of synapses with a 
single neuron at their centre are affected in a similar way at the same time, but 
without fully removing the synapses or any neurons from the resulting output
patterns used for evaluating network performance.

With this in mind, neurofibrillary tangles (NFTs) of hyperphosphorylated tau are 
known to result in direct blocking of axonal transport \cite{ballatore2007tau} 
and collapse of microtubules also supporting axonal transport
\cite{feinstein2005inability}. In order to model more specifically the 
pathological effects of tau NFTs it is suggested that, rather than simply 
deleting neurons or connections at random or in areas of a certain radius
\cite{ruppin1995neuralimpairment, ruppin1995patterns}, the output of selected 
neurons could be partially muted to simulate the effects of axonal blocking by 
NFTs.

To simulate the sequestering of hyperphosphorylated tau and the subsequent 
cascading spread of damage, neighbouring neurons could also be muted by a 
slightly smaller amount, with new tau lesion centres subsequently formed near to 
existing lesions, and the resulting distributed damage occurring in less of a 
severe `binary' manner as with random deletion of synapses or neurons.

\begin{figure}[t]
	\centering
	\includegraphics[width=0.48\textwidth]{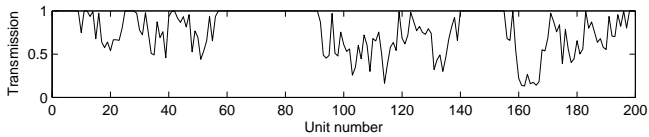}
	\caption{Characteristic plot of neural damping levels after spatial tau
	lesioning.}
	\label{fig:taulesioning}
\end{figure}

In computational terms, lesioning can be performed in steps of size $z$. The
locations of the centres of the first set of lesions (the tau \textit{seeds}) are 
chosen at flat random from across all neurons and are assigned to set $D$. 
Subsequent lesioning steps proceed as follows:

\begin{itemize}
  \item A subset $d \subseteq D$ of size $z$ locations are chosen at flat random. 
  With Gaussian probability centred on each element in $d$, a new set $d'$ of 
  neighbours is chosen.
  \item Each neuron's activation is dampened by multiplying by a value drawn from
  a Gaussian distribution as a function of the neuron's proximity to the lesion 
  centre (i.e. $x-\mu$), and with lesion width $\sigma$ (suggested as $2$),
  such that those neurons closest to the lesion centres in $d'$ are most heavily
  diluted, and those distant from the elements of $d'$ are relatively unaffected.
  \item $d'$ is added to $D$ and the process is repeated with the new, larger set
  of lesion centres $D$. This results in characteristic lesioning as seen in
  figure \ref{fig:taulesioning}.
\end{itemize}

\section{Results}
\label{results}
Unless otherwise stated, all experiments were performed in a network with the
following parameters: network size $N = 1600$, connections per unit $K=200$,
neural threshold $\theta = 0.048$, noise $T = 0.005$, learning rate $\gamma =
0.025$, external input strength (learning mode) $e_l = 0.065$, external input
strength (retrieval mode) $e_r = 0.035$, coding rate $p = 0.1$, deletion step
$\Delta d = 0.01$. Results were averaged over 10 runs and the number of
patterns stored on each run was 10.

\subsection{Random deletion and local field-dependent compensation}
\begin{figure}[t]
	\centering
	\includegraphics[width=0.48\textwidth]{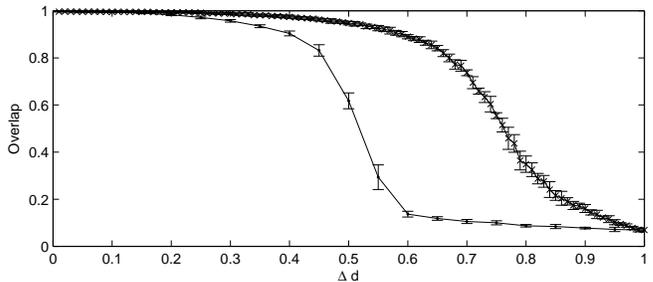}
	\caption{Performance over synaptic deletion without compensation (leftmost
	curve) and with (rightmost curve). The deletion step size $\Delta d$ is
	larger for the curve without compensation, but this does not affect the
	overall result as deletion step size-sensitivity is only introduced with
	compensation.}
	\label{fig:compensation}
\end{figure}

In the first experiment, an attempt was made to replicate the results of Ruppin
and Reggia \cite{ruppin1995neuralimpairment} using the improved neural
field-dependent compensation rule of Horn et al. \cite{horn1996neuronal}, in
which compensatory mechanisms extended the working life of the network during
repeated synaptic deletion. A set of 10 patterns was stored in the network and
the average retrieval success rate (\textit{overlap}) after various levels of
deletion was plotted. The results shown in figure \ref{fig:compensation} are
comparable with those achieved by Horn et al. \cite{horn1996neuronal} and can be
used as a performance baseline for later experiments. The network was connected
with a Gaussian connection strategy.

\subsection{Compensation using recent versus remote memories}
\label{compensationresults}
\begin{figure}[t]
	\centering
	\subfloat[Final deletion = 0.35]
	{\label{fig:0.35compsets}\includegraphics[width=0.23\textwidth,
	trim=0 0 0 20,clip=true]{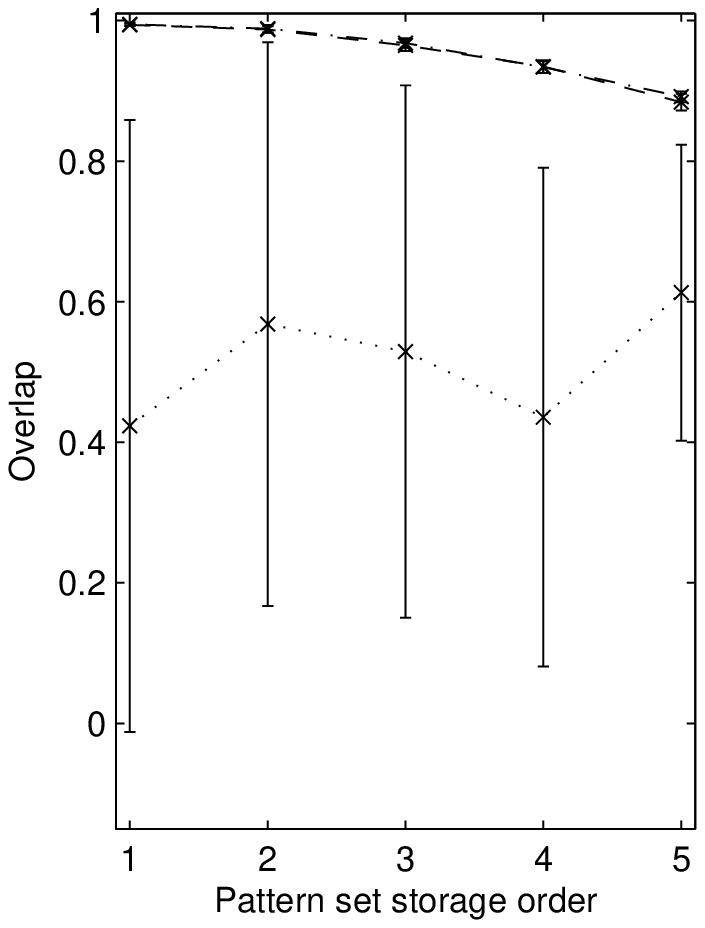}}
	\subfloat[Final deletion = 0.45]
	{\label{fig:0.45compsets}\includegraphics[width=0.23\textwidth,trim=0 0 0
	20,clip=true]{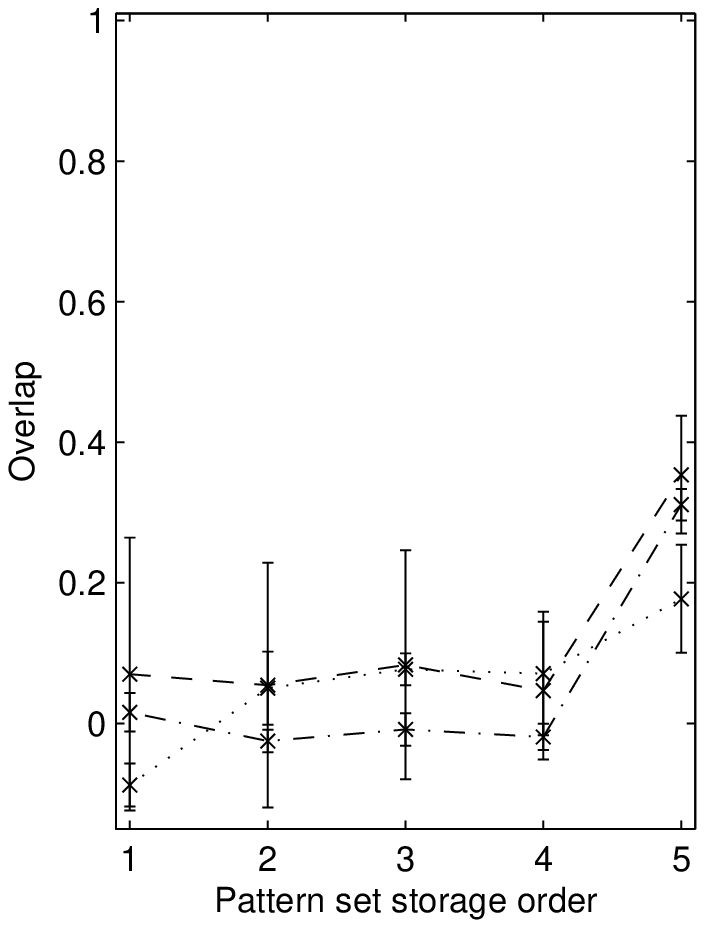}}
	
	\caption{Performance on
	separately stored sets of memories. The network was alternately presented with
	sets of 6 patterns then subjected to a process of deletion with
	compensation using only the first set of patterns stored (dotted line), only
	the last set of patterns stored (dot-dash line), or using a random set of 6
	patterns drawn from all those previously stored (dashed line). By the
	final round, the total proportion of deletion was either 0.35 (fig
	\ref{fig:0.35compsets}) or 0.45 (fig \ref{fig:0.45compsets}).}
	
	\label{fig:compensationsets}
\end{figure}

Ruppin and Reggia observed a gradient of damage by repeating a process of
learning a set of patterns then subsequently deleting a proportion of the
connections between units \cite{ruppin1995neuralimpairment}. Their results,
based on a fixed synaptic compensation strategy, showed a clear decrease in
recall performance for patterns learned recently compared with those learned
earlier in the process. These results were replicated, and are shown in figure
\ref{fig:compensationsets}.

As the local field-dependent compensation strategy of Horn et al.
\cite{horn1996neuronal} works by using the retrieval of stored memories for
comparing average post-synaptic potentials before and after damage, the choice
of memories which should be used for this purpose becomes significant due to the
different retrieval success rates of patterns stored early in the lesioning
process compared to those stored more recently.

 \begin{figure}[t]
    \centering
    \includegraphics[width=0.48\textwidth]{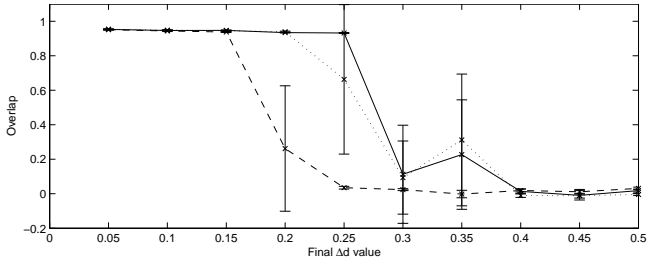}
    \caption{Average performance on all learned sets (5 sets of 10 patterns,
    $N=1200$, $K=200$) after different final levels of deletion. First-set
    compensation (dashed line) performance declines significantly earlier than
    latest-set compensation (dotted line), which itself is marginally less robust
    than random-set compensation (solid line).}
	\label{fig:delgradbatch}
\end{figure}

As shown in figure \ref{fig:0.35compsets}, if only remotely-stored patterns are
used during the compensatory process (dotted line) the performance of the
network is severely degraded even at a relatively low level of deletion, whilst
use of the most recently-stored patterns during compensation is almost
indistinguishable in performance from the results when using a random set of
patterns drawn from all those previously stored (dot-dash and dashed lines,
respectively).

In these cases, a clear gradient of learning has been observed such that
patterns stored most remotely are recalled more successfully than those stored
more recently (note that this is a distinct phenomenon from serial-position
effects in which recency and primacy of items within a list correlate with
greater recall, as the network is storing time-separated sets of patterns in
between periods of damage rather than a single list of items). At higher,
catastrophic levels of deletion (figure \ref{fig:0.45compsets}), compensation
using randomly-selected patterns slightly outperforms compensation using only
the most recently-stored patterns, but within the margins of error.

Despite the greater accuracy of their recall within a functioning network, using
only remotely-stored patterns during compensation results in much earlier decline
of the network performance as deletion progresses (figure
\ref{fig:delgradbatch}). Conversely, compensation using the most recently-stored
patterns, or sets of patterns drawn at random, results in greater robustness to
damage. This could be due to the effect of decreased variance in the patterns
used to calculate the signal term during compensation when using only a small,
fixed set of patterns, resulting in increased noise during the compensatory
process. Any noise arising from using only the first learned set during
compensation is multiplied on each compensatory step due to the lower input
variance. When using the latest set of stored patterns or a random set at each
compensatory step, the variance of the data is increased and this helps to keep
noise to a minimum.

This has implications for sufferers of Alzheimer's disease. If the network is
damaged to such an extent that the gradient observed in figure
\ref{fig:0.35compsets} (the uppermost dashed and dot-dash lines) is evident, but
the network is not yet catastrophically damaged, then there may be a greater
likelihood that compensatory mechanisms will use remotely-stored patterns
compared to recently-stored ones due to their higher recall success. As this
experiment has shown, this could actually lead to earlier overall decline of
cognitive abilities, and a cycle of correspondingly worse recent memory
retrieval performance and thus continued use of remote memories during
compensation. Additionally, this finding places the deletion threshold of the
model (beyond which all recall is severely affected) closer to the reported
10-30\% atrophy levels seen before the symptomatic damage evident in AD.

\subsection{Connection strategies}
\subsubsection{Effects on network capacity}
\begin{figure}[t]
    \centering
    \includegraphics[width=0.48\textwidth]{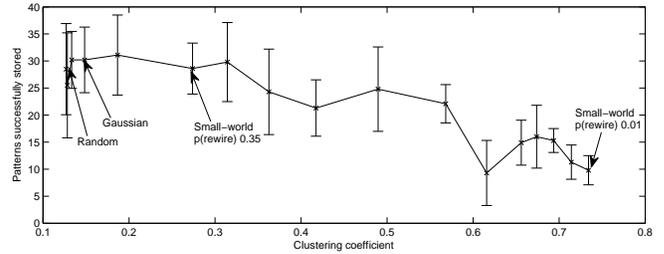}
    \caption{Capacity in networks with different small-world clustering
    coefficients. Flat-random, Gaussian and representative
    small-world networks are indicated individually.}
	\label{fig:capacityconn}
\end{figure}

Next, the effects on network capacity and robustness to damage of various
connection strategies were compared. Firstly, networks were created with $N=800$
units with connection density $K=0.125N$. The networks were wired with Gaussian,
flat-random, and small-world (with various values for $p(rewire)$) connectivity.
Patterns were stored in each network according to equation \ref{eq:actdep} and
retrieved immediately after storage. The average retrieval success rate was
plotted against the small-world clustering coefficient of the network's
connection matrix, and when the average overlap measure dropped continuously
below $0.8$, the network was assumed to have reached its capacity (figure
\ref{fig:capacityconn}).

\begin{figure}[t]
    \centering
    \includegraphics[width=0.48\textwidth]{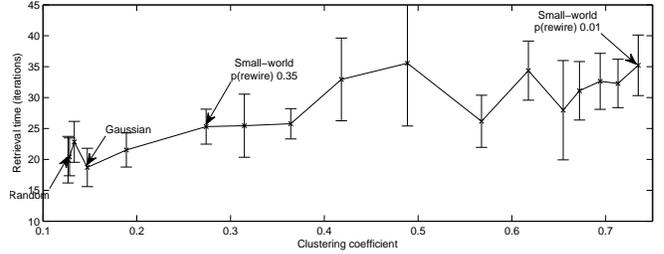}
    \caption{Retrieval times in networks with different small-world clustering
    coefficients.}
	\label{fig:retrievaltimes}
\end{figure}

\begin{figure}[t]
	\centering
 	\subfloat[Small-world (0.01)]
 	{\label{fig:smallworld0.01conn}\includegraphics[width=0.23\textwidth]{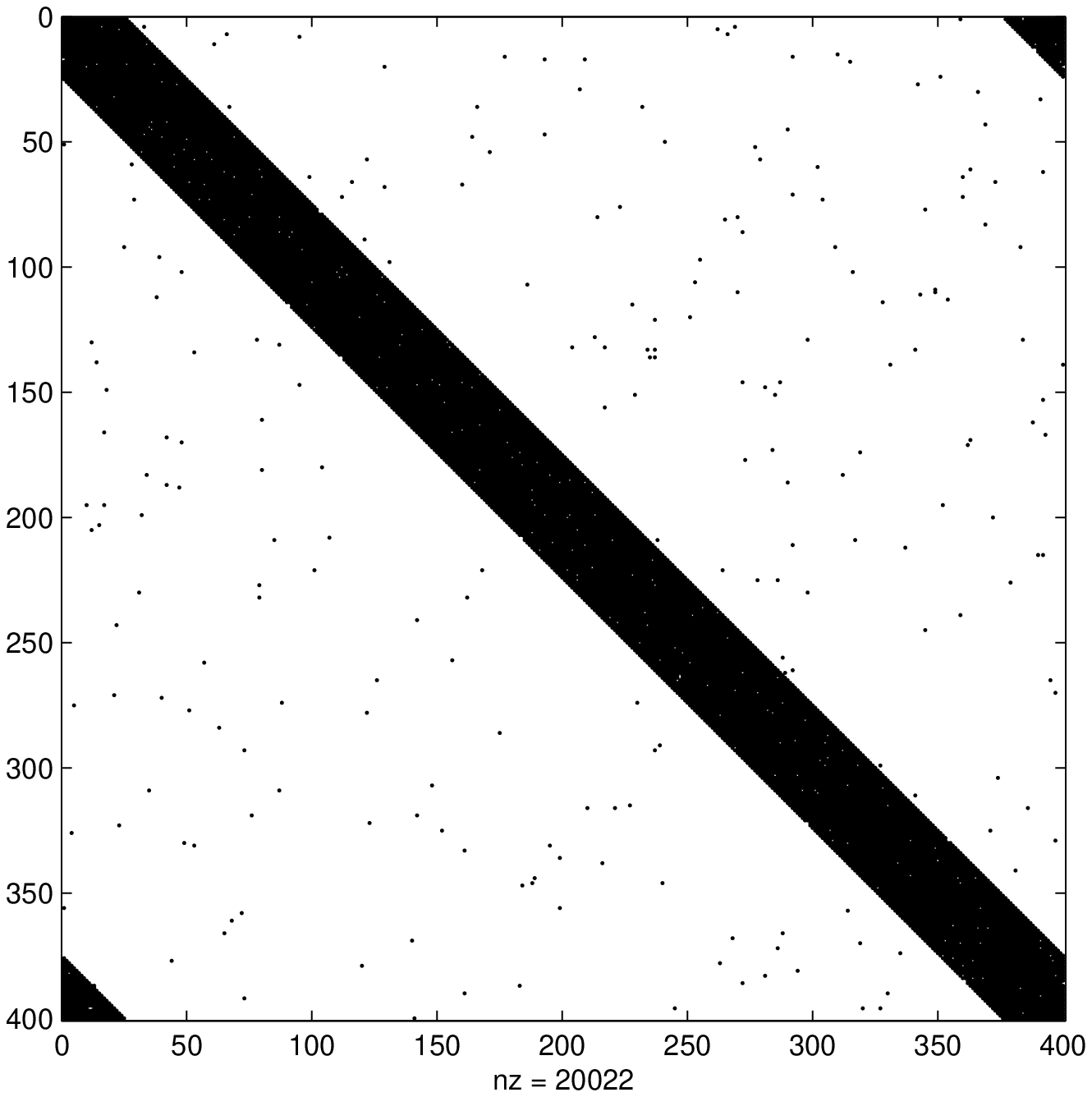}}
 	\subfloat[Small-world (0.9)]
 	{\label{fig:smallworld0.9conn}\includegraphics[width=0.23\textwidth]{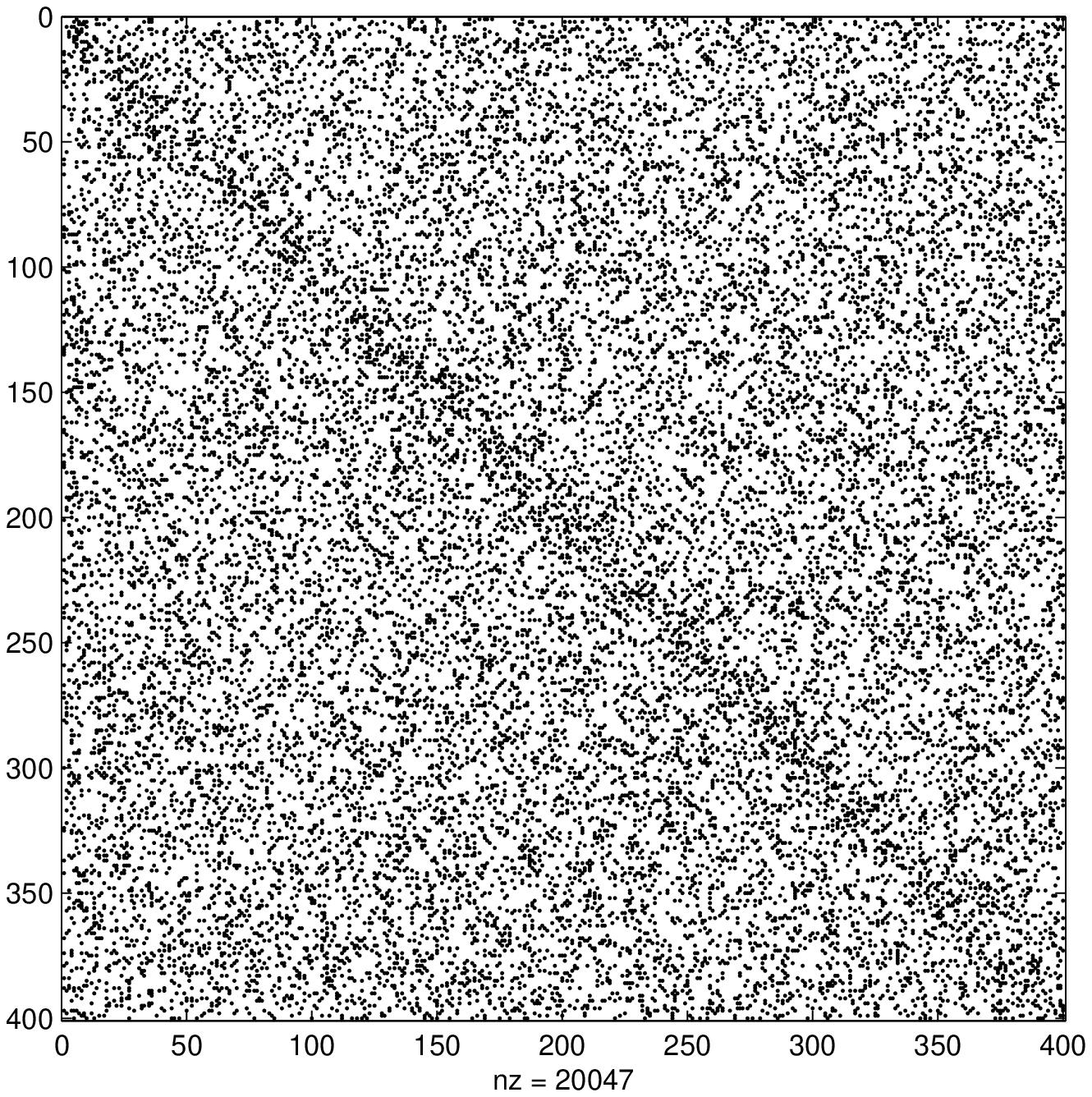}}
 	\caption{Connection matrices of networks connected in a small-world manner
	($N=1600$, $K=200$, $p(rewire)=0.01, 0.9$, only every fourth connection
	plotted for clarity).}
	\label{fig:connectionmatrices}
\end{figure}

The results indicate that specific network connectivity generally has little
effect on capacity, with the random and Gaussian networks appearing on the same
trend as the small-world networks, but it appears that network capacity is
significantly reduced in more highly-ordered networks with high clustering
coefficients such as small-world networks with low values for $p(rewire)$.
Although each network structure contains the same number of connections as the
others and should therefore have effectively the same capacity, the difference
becomes clear when examining the layout of the connections and the related
small-world clustering coefficients. Compare figures
\ref{fig:smallworld0.01conn} and \ref{fig:smallworld0.9conn}, both of which were
constructed using the small-world algorithm. The majority of connections in the
network with $p(rewire)=0.01$ are located incredibly densely within the local
neighbourhood of each neuron, with only a few projections to more distant parts
of the network, resulting in a clustering coefficient of 0.73.

This appears to have two effects: the first becomes clear when considering the
pattern recall times in figure \ref{fig:retrievaltimes}, which shows that in
highly-regular networks the number of iterations required for the network to
fall into a stable state is higher. This is likely to be due to the lack of
distant projections to other parts of the network: activation is `slowed-down'
by having to flow through a closely-linked chain of units from one extent of the
network to the other, whilst a network with less regularity and more distant
projections (as seen in figure \ref{fig:smallworld0.9conn}) can effectively take
short-cuts when activation to distant parts of the network is required. If this
activation degrades over time as it traverses the network in small steps, or if
there is a limit to the permitted time between cueing and retrieval of a
pattern, it is clear to see that these effects could result in greater retrieval
failure rates (and thus lower effective capacity) than in a network with less
regularity in its connection matrix.

\subsubsection{Effects on redundancy and robustness}
The second effect concerns the information capacity of the connections in the
network. The high density of local connections in the regular network leads to
synaptic redundancy, as activation between any two nearby neurons can take
multiple paths between them. Necessarily, redundancy where more than one
connection carries the same information results in a reduction in information
capacity elsewhere in the network, as previously shown, but increased redundancy
should also lead to networks which are more robust to damage.

\begin{figure}[t]
    \centering
    \includegraphics[width=0.48\textwidth]{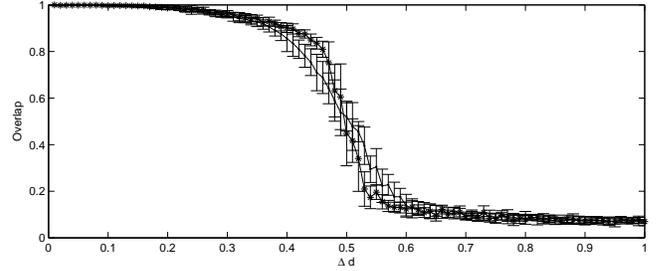}
    \caption{Deletion without compensation in random (star
    markers) and highly-ordered (dot markers) networks.}
	\label{fig:infocapacity}
\end{figure}

To test this prediction, a profile of deletion (without compensation) was
obtained for networks connected with small-world ($p(rewire)=0.01$) and
flat-random connectivities in 1600-unit networks with connectivity $K=0.125N$.
The resulting plot in figure \ref{fig:infocapacity} shows a marginally smoother
rate of decline and greater longevity of performance in the small-world network
(dot markers) than in the random network (star markers), indicating that the
high local connectivity density does indeed lead to redundancy and hence greater
robustness to damage, but at the expense of lower capacity. Nevertheless, the
effects are relatively small overall.

\subsection{Tau lesioning}
\begin{figure}[t]
	\centering
	\subfloat[Standard-rate tau lesioning]
 	{\label{fig:standardtau}\includegraphics[width=0.48\textwidth]{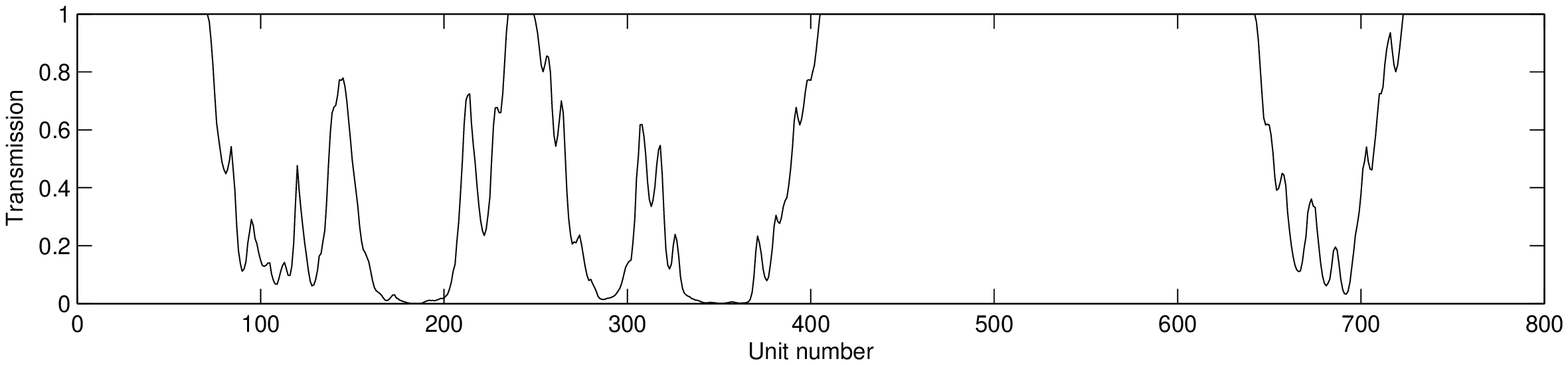}}
 	\\
 	\subfloat[Enhanced-rate tau lesioning]
 	{\label{fig:acceleratedtau}\includegraphics[width=0.48\textwidth]{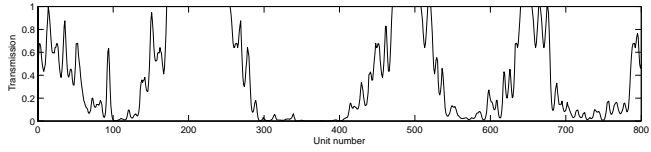}}
 	\caption{Lesions applied to example networks after every unit has been
 	considered once. In the second graph, the rate of reduction of neural
 	transmission in each lesioning step has been squared and the horizontal
 	spreading speed of the lesioning has been doubled.}
	\label{fig:taulesionexamples}
\end{figure}

To identify the changes in behaviour when more distributed, variable-rate tau
damage occurs within the network, a network with $N=800$ units was connected in
a Gaussian manner and tau lesioning was performed according to the method
described in section \ref{tauimplementation}, with random-set compensation. Two
rates of tau lesioning were inspected: in addition to the standard rate in which
the neuronal outputs were muted by an inverse Gaussian probability as a function
of the neuron's distance from the lesion centre, a second rate was tested in
which the muting amount was squared so as to increase the speed with which the
lesions resulted in full neuronal blocking, and the width of the distribution
used for choosing new nearby tau lesion centres was doubled. Examples of the
resulting comparable increase in lesioning can be seen in figure
\ref{fig:taulesionexamples}. A further, currently untested, method of altering
the tau lesioning rate would be to consider each unit more than once until full
blocking of all units occurs.

\begin{figure}[t]
	\centering
	\includegraphics[width=0.48\textwidth]{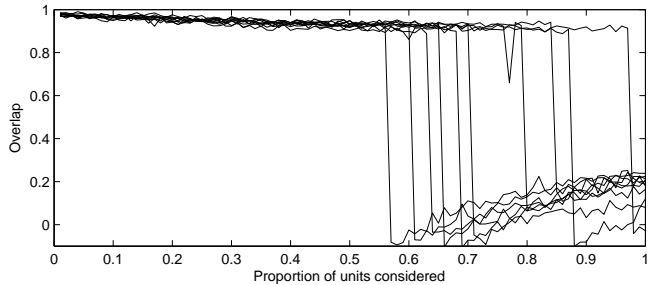}
	\caption{Performance plotted against number of units considered for lesioning 
	with tau (enhanced rate). Due to the large differences in the time of onset of
	impairment, individual test runs have been plotted for easier comparison.}
	\label{fig:taulesionprofiles}
\end{figure}

The results in figure \ref{fig:taulesionprofiles} show a very different profile
to basic deletion (see figure \ref{fig:compensation}). Rather than a smooth
decline in performance which tails off towards zero, a sudden catastrophic
decline in performance occurs during a single step of tau lesioning. The
performance then steadily recovers, but only to a fraction of the original
performance, as the compensatory mechanisms attempt to ``catch up'' with the
sudden decrease in activation. As seen in figure \ref{fig:taulesionexamples},
there are still areas of the network which are undamaged (transmission remains
at 1), and it is likely that it is these areas which contribute to the
above-zero final performance of the network.

Although the precise timing of the sudden decline varies randomly between test
runs, each line on the graph traces essentially the same shape. Indeed, it was
found that the differing rates of neuronal damping shown in figure
\ref{fig:taulesionexamples} resulted in exactly the same profile of altered
performance during lesioning, except that the performance drop-off was
experienced correspondingly earlier or later with faster or slower tau lesioning
rates.

\begin{figure}[t]
	\centering
	\includegraphics[width=0.48\textwidth]{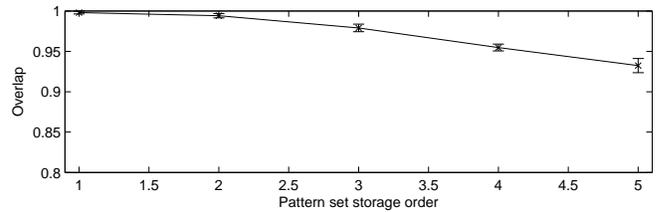}
	\caption{Gradient of performance on sets of patterns stored at different
	stages of lesioning with tau.}
	\label{fig:taugradient}
\end{figure}

To test that the observed gradient of learning in sets of patterns over time
(figure \ref{fig:compensationsets}) still occurs with tau lesioning, the
experiment in section \ref{compensationresults} was re-run in a
Gaussian-connected network with tau lesioning instead of deletion. The patterns
used for compensation were drawn at random from all those previously stored. The
results in figure \ref{fig:taugradient} are comparable with those in figure
\ref{fig:compensationsets}, indicating that tau lesioning in this way does not
destroy the effect of reduced retrieval of recent compared to remote patterns.

\section{Conclusions and further work}
\label{conclusions}
This work has presented updates to a long-standing and widely-cited
computational model of Alzheimer's Disease \cite{ruppin1995neuralimpairment},
including first successfully replicating, and then extending, the experiments of
Horn et al. \cite{horn1996neuronal} on the effects of local, field-dependent
synaptic compensatory mechanisms within the model.

The differing effects of using recent, remote, and random sets of memories to
calculate compensatory signal terms has been shown, revealing that the network
is sensitive to the choice of which set is used. Using only remote memories to
calculate the signal term results in greater noise within the compensatory
mechanism, and an earlier decline in performance as synapses are deleted (much
closer to the $10-30\%$ range seen in AD patients \cite{minati2009reviews}). The
implications for AD patients are shown in the context that initial retrieval of
remote memories at early stages of damage is actually more reliable than with
recent memories: if the brain makes use of this effect and uses the more
readily-available remote memories to calculate compensation, not only do the
recently-stored memories continue to become less reliable than the remote
memories, but the noise in the system leads to earlier onset of catastrophic
decline.

Speculatively, if the biological realisation of synaptic compensation via memory
retrieval could be considered as dreaming (as postulated by Horn et al.
\cite{horn1996neuronal}), these results are consistent with the idea that the
greatest compensatory success is likely to be found by using memories and states
acquired throughout an individual's lifetime in the compensatory mechanism, or
by using those memories most recently obtained, rather than primarily memories
from early life. Further studies to examine any potential link between this
effect and any reported fixation during dreaming on remote memories in AD
patients (either prior to, or after, onset of symptoms) could yield important
results.

It has also been shown that network capacity and resilience is related to the
regularity of connections within the network. High small-world clustering
coefficients lead to redundancy within the network, meaning greater resilience
to damage but at the expense of lower capacity, as well as longer pattern
retrieval times. This is consistent with the findings of Supekar et al.
\cite{supekar2008network} who examined small-world functional networks in the
brain and found a key correlation between loss of small-world connectivity and
onset of AD symptoms. Further examination of the relationships between
small-world clustering, robustness, retrieval speed and network capacity could
be revealing, as well as studies into how this operates within the principle of
neural Darwinism (pruning of weaker synapses during brain development).

Lesioning with simulated tau rather than standard synaptic deletion has been
shown to create a very different profile of damage by allowing all neurons and
synaptic connections to remain present (so output patterns are not artificially
altered) and instead damping inter-neuronal transmission. Whilst initially
offering a much more graceful decline in performance due to the persistence of
synaptic connections and output units, consistent with the slow degradation seen
in AD, the drop-off in performance when it finally occurs is much more severe
with tau lesioning than with synaptic deletion despite some later compensatory
recovery of performance.

Further work will be needed to ascertain whether this deletion profile offers a
more plausible explanation of AD symptoms and whether the observed temporary
improvement in recall after some level of catastrophic damage can be medically
corroborated, but it must be borne in mind that tau pathology represents only a
subset of the processes underlying AD. It would be beneficial to extend this
concept and show in a similar way the effects of alternative AD pathologies. Of
particular interest are the beta-amyloid mechanism and its extension, the N-APP
hypothesis, in which a fragment of the amyloid precursor protein (N-APP) is
found to be capable of binding to the \textit{DR6} cell-death receptors of
neuronal cell bodies and axons, with the effect of accelerating apoptopic cell
death. The apoptopic mechanism involves the release of \textit{caspases}, of
which caspase 6 is capable of cleaving the N-APP fragment from existing
$\beta$-amyloid deposits, leading to a  cascade of neurodegeneration
\cite{nicholson2009neuroscience}.

The Tsodyks and Feigel'Man model studied in this work is only a basic
associative network with limitations in processing ability and a relatively
constrained range of behaviour. More sophisticated artificial neural
network-based models such as LEABRA \cite{oreilly2001generalization}, spiking
neurons \cite{gerstner2002spiking}, and reservoir networks
\cite{lukosevicius2009reservoir} are available and could provide further
insights into the effects highlighted in this paper. In particular, it would be
interesting to examine the effects of synaptic compensation and connectivity
strategies within a reservoir computing framework due to the large potential for
exploration of the currently poorly-understood dynamics, and the greater
computational power (potentially offering the representation of more varied
symptoms of AD than simple pattern recall) of these systems.

\section{Acknowledgement}
With thanks to my academic supervisor, Dr. John Bullinaria, for his comments.

\bibliographystyle{plain}
\bibliography{bibliography}
\end{document}